# EPIK: Eliminating multi-model Pipelines with Knowledge-distillation

Bhavesh Laddagiri[1], Yash Raj[2], and Anshuman Dash[3]

¹SRM Institute of Science and Technology
²Chandigarh University

*Abstract*—**Real-world tasks are largely composed of multiple models, each performing a sub-task in a larger chain of tasks, i.e., using the output from a model as input for another model in a multi-model pipeline. A model like MATRa performs the task of Crosslingual Transliteration in two stages, using English as an intermediate transliteration target when transliterating between two indic languages. We propose a novel distillation technique, EPIK, that condenses two-stage pipelines for hierarchical tasks into a single end-to-end model without compromising performance. This method can create end-to-end models for tasks without needing a dedicated end-to-end dataset, solving the data scarcity problem. The EPIK model has been distilled from the MATRa model using this technique of knowledge distillation. The MATRa model can perform crosslingual transliteration between 5 languages - English, Hindi, Tamil, Kannada and Bengali. The EPIK model executes the task of transliteration without any intermediate English output while retaining the performance and accuracy of the MATRa model. The EPIK model can perform transliteration with an average CER score of 0.015 and average phonetic accuracy of 92.1%. In addition, the average time for execution has reduced by 54.3% as compared to the teacher model and has a similarity score of 97.5% with the teacher encoder. In a few cases, the EPIK model (student model) can outperform the MATra model (teacher model) even though it has been distilled from the MATra model.**

## I. INTRODUCTION

**K**NOWLEDGE DISTILLATION [1] is a process of transferring knowledge from a source model, known as the teacher, to a new model, known as the student. The goal of this process is for the student model to replicate the teacher model as closely as possible, including replicating the teacher's correct and incorrect predictions. For accurate results from the student, the teacher model should perform optimally, making as few errors as possible.

The distillation process has existed in many forms and is done due to several reasons like reduction of model size or time of execution. Models like DistilBERT [2] have been distilled from BERT [3] mainly because the size of BERT was enormous, and distil-BERT could retain 97% of the performance by eliminating 50% of the parameters.

The authors of this paper propose the EPIK model, which has been created by distilling the knowledge of two models to perform tasks for which direct datasets do not exist. This technique of knowledge distillation has been developed to reduce the chaining process (where the output of the first model is used as input to the subsequent models in a chain manner) into a single model. Currently, no similar systems are available that can create such models from existing models. Instead, every available solution which proposes the conversion of chained approaches to direct models demands the generation of datasets on which models could be trained by either gathering new data or merging existing data. This process of knowledge distillation proposes a new method that depends on already trained models and removes the dependence on new datasets.

For example, there have been no deep learning models for direct transliteration of several Indian languages to other Indian languages, like transliteration of text from Bengali to Hindi or Kannada to Tamil. In recent works, the MATRa model [4] was proposed, which used English as the intermediate output to transliterate from a source Indian language to any target Indian language. There are no publicly available datasets for direct transliteration between any pair, and the only publicly available datasets have English as either the input or output word. Thus, the transliteration system had to be executed in two steps where the source text was input in any Indian language (like Bengali) and the model predicted the corresponding English transliterated text. This English output was given to the model, which predicted the corresponding transliterated text in the required target language (like Hindi). In the practical sense, several non-mainstream tasks, such as Indic to Indic transliteration, lack datasets to complete the task in a single step. Thus, building direct solutions without datasets is impossible, and a multi-step chaining approach is applied wherever possible.

The models generated using this method of knowledge distillation could be a distilled version of two or more models to accomplish any task in a single step. In this paper, the authors have demonstrated the distillation of the MATRa model [4] into a single transformer model that can execute Indic-to-Indic transliteration without the need for a dataset. The MATRa model is a combination of 2 separate multilingual models - an Indic-to-English transliteration model and an English-to-Indic transliteration model. Thus, MATRa had to use English as the intermediate output for Indic-to-Indic transliteration. On the contrary, the EPIK model can do the direct tasks of Indic-to-Indic transliteration by eliminating the English intermediate output. It has been distilled from the two models in the MATRa system and can replicate the teacher model in 99% of all the cases. There are a few instances when the teacher (MATra model) makes an error, but the student (EPIK model) doesn't make that same error.

This technique of knowledge distillation is not limited to translation-like tasks, it can be applied to several other tasks like Computer Vision or reinforcement learning (section III). Furthermore, the student and teacher model can have separate architectures. This knowledge distillation method can work for different architectures like LSTM-based models distilled to transformer-based models. Overall, the EPIK model combines several separate teacher models distilled into a single student model, with negligible reduction in performance metrics, with some emergent results where it outperforms the teacher in a few instances.

## II. MOTIVATION

### A. Issues in MATra model (teacher model)

The MATra model could transliterate from a source Indic language to a target Indic language (among Hindi, Bengali, Tamil or Kannada). However, it was done in two stages, with English as the intermediate output. This was mainly because there were no publicly available datasets for direct transliteration from any Indic to any other Indic language. The only available dataset had English as either the source or target word. Thus, two multilingual models were built and combined to form the MATra model. This issue has existed for a long time for several non-mainstream tasks where the required dataset has not existed, resulting in the development of such two-staged or multi-staged approaches. To execute the required tasks (with no direct datasets), separate models have to be developed with a common intermediate output creating a chain of outputs for the final result. There are several issues, such as -

1) The total time for execution was dependent on the individual execution time of each model

2) The total memory consumed was a summation of the memory required to run each model separately in addition to storing the intermediate outputs at each stage

3) The total errors were added at each stage, and the final output would not be fully accurate if any of the models made a small mistake. If the first model in the chain made the error, all subsequent models would magnify the error, and the final output would be completely incorrect.

### B. Issues in merging or creating different datasets

Regarding translation in Indian languages, there are over 270 spoken languages, and Google translate has only 19 languages available for translation. Furthermore, no major translation system apart from Google translate is available for translating text between language pairs like English and any remaining Indic language. If we wanted to build any translation system for translation tasks among all these languages, we would have to create datasets for all required pairs because we do not want a 2-step-translation with intermediate outputs. For the translation of all 270 languages to each other, we would need $36,585$ datasets that map each language to the other. With only 100,000 sentences in each dataset, we would have a total corpus size of $36,585 * 100,000$ sentences. Considering the size of the entire corpus, it is clearly hard even to generate such a corpus, and training such models would require a lot of resources.

Another approach to accommodate all possible combinations could be to keep a constant set of source texts in a single primary language (like Hindi) and have the corresponding mapping in the required target language (any of the remaining 269 languages). When we need the translation for any language pair which does not involve the primary language, we can do the required mapping (using the primary language as the intermediate language) and generate the training dataset dynamically in the required pair format. This way, we can generate any dataset from the entire corpus of $36,585$ datasets only when the requirement arises. The major issue with this approach is that the training corpus would be limited to the constant set of source texts. It will have minimal variation and diversity. Therefore, while predicting any sentence outside the training dataset, it would struggle to generate accurate results. In addition, the total memory required would naturally be the same as the earlier example (where we have the entire corpus of $36,585 * 100,000$ sentences), with the difference that any pair could be loaded at the required time only, and flexibility is higher in this format.

This technique of knowledge distillation (used by the EPIK model) requires two or more teacher models that adopt a chain approach (where the output of the first model is used as input to the subsequent models in a chain manner) to generate the final output. However, with the availability of all the source models, we could generate the required dataset without the intermediate outputs and then create a basic model that could do the required task directly instead of distilling knowledge into the student model. For example, the MATra model has already been trained to perform Indic to Indic transliteration. It has an accuracy score of 93.5%, but does the process in 2 steps using English as the intermediate output. With the availability and access to the MATra model, we could generate a dataset of any required size using the MATra model and train the EPIK model on the generated dataset instead of distilling knowledge. The authors tried to solve the task of direct transliteration using this format of dataset generation, but the final outputs had significant degradation in accuracy and acceptability (about 8.7% reduction). The primary issue is that the MATra model was trained on a pure dataset with perfectly transliterated data but could achieve a maximum accuracy of only 93.5%. Thus, the purity of the dataset generated by the MATra model would only be 93.5%, and the new student model had an average accuracy of only 84.8% when trained on the generated dataset.

In contrast, the EPIK model achieves an accuracy score of 92.1% (with a reduction of only 1.4%) using this format of knowledge distillation. Considering other issues like the time required to generate the direct dataset was extremely high, the total size of the generated dataset grew by a large factor. So the approach of dataset generation was eliminated.

The EPIK model can easily overcome such issues by distilling the knowledge of several models without the need for a huge dataset or high computation resources. In addition to generic tasks, this process can be used to execute tasks such as image captioning in Sanskrit (further discussed in section III) or other tasks that do not have a dataset or the generation of datasets is a computationally expensive task.

## III. Applications

The EPIK model is a student model that has been distilled from 2 or more teacher models and is able to execute all the tasks of the individual teacher models in a single step. Some practical uses of such distillation techniques are -

1) Image Captioning in any low-resource language (like Sanskrit): To execute the direct task of image captioning in low-resource languages (like Sanskrit), very few systems have been built to date, mostly due to the lack of datasets. Several models exist for the task of image captioning in English and the task of translation from English to other low-resource languages (like Sanskrit). However, suppose any system exists for this task. In that case, it is done using a chaining mechanism. The image is first captioned in English, and the intermediate sentence is translated into the target language (like Sanskrit).

   However, the models generated with this technique of knowledge distillation can easily overcome such issues of a multi-staged approach even without datasets. The two separate existing models can be used to create a single student model where the encodings of the student model will be able to directly match the encodings of the captioned English sentence (these encodings are from the second model), and the decoder will be fine-tuned to give a caption in the target language.

2) Conversion of architectures for existing tasks (any encoder-decoder-based model): If the current approach is using an encoder-decoder-based model (like LSTM), it can easily be distilled into any other encoder-decoder-based model (like transformers) using this method of knowledge distillation. For example, most NLP tasks had models built using LSTMs until the introduction of transformers in 2017. However, there was a sudden shift to transformer-based models in most tasks. The only issue is that transformer-based models need a larger dataset than an LSTM-based model, and thus, the size of datasets for each task grew by a large factor.

   However, in this knowledge distillation approach, the student model's encoder (like a transformer [5]) can be directly trained to match the encoder of the teacher model (like LSTM). The decoder (of the student model) can then be trained to give the output depending on the target. The major advantage of such a method of transferring knowledge is that the model's architecture can be shifted completely without a change in the dataset or a reduction in the performance of the student model. In general, transformers need a larger dataset to give equivalent performance to LSTM models for the same task. However, in this method, during the conversion of architectures, the student model needs to match the teacher model. Thus, it just required the original dataset without any addition of data.

3) Translation of low-resource languages to other low-resource languages: As elaborated in the motivation section (section II), there exist 270 spoken languages in India. If we have to create a direct multi-lingual translation model for all possible language pairs, we must have all combinations of datasets available. Overall, we would need 36, 585 datasets that map each language to the other; it is impossible to generate such datasets. In addition to this problem, no two people may know a certain pair of languages (for example, no one may know both Assamese and Tamil). The creation of such pairs would be impossible.

   Thus, using this technique of knowledge distillation, we need to create 270 separate translation models that map each language to a common primary language (each model can be trained on different datasets from different sources, having no common data samples between them), and direct translation between all 270 languages is possible. Furthermore, this method of knowledge distillation can create a translation system between language pairs where the creation of a dataset is impossible.

Apart from these practical uses, there are several advantages to distilling knowledge into a student model. Some examples are -

1) The student model takes lesser memory (as it will be running only one student model instead of two or more teacher models).

2) The execution time of the student model will be lesser than the combined time of respective teacher models.

3) The chances of errors are reduced. In any system that depends on the chaining of models, each model must perform perfectly without errors. However, the student model is a generalised version of the chain and is not sensitive to the noise or errors caused by a single model in the whole chain. Thus, the chances of errors are nullified after distillation. (section VI)

## IV. Related Work

*1) Knowledge Distillation:* Distillation is a process that is used to compress larger models while retaining performance. DistilBERT [2] is a model from HuggingFace that compressed the original BERT [3] model. DistilBERT reduced the size of the BERT [3] model by 40% while retaining 97% of its language understanding capabilities and being 60% faster.

The paper KD-Net [6] is a framework to transfer knowledge from a trained multi-modal network (teacher) to a mono-modal one (student). This paper demonstrates a new distillation framework where the student network is trained on a subset (1 modality) of the teacher's inputs (n modalities).

Along similar lines, the paper Cross Modality Knowledge Distillation (CMKD) [7] presents 2 models which leverage semi-supervised enhanced training and mutually transfer knowledge to strengthen the model. The paper

"Multilingual Neural Machine Translation with Knowledge Distillation" [8] shows that it is possible to execute the task of multi-lingual translation by first training different models and then distilling them to produce almost similar results.

*2) Multi-Staged approach:* In the paper "Two-Stage Textual Knowledge Distillation for End-to-End Spoken Language Understanding" [9], the authors propose a two-stage textual method that uses vqwav2vec BERT as a speech encoder which helps to improve the performance, especially in a low-resource scenario, for the task of efficient spoken language understanding in a pipeline manner.

The paper "M2KD: Multi-model and Multi-level Knowledge Distillation for Incremental Learning" [10] proposes a multi-model and multi-level knowledge distillation strategy where they directly leverage all previous model snapshots instead of sequentially distilling knowledge only from the last model. In addition, they incorporate an auxiliary distillation method to further preserve knowledge encoded at the intermediate feature levels.

The authors of the paper "Model Compression with Two-stage Multi-teacher Knowledge Distillation for Web Question Answering System" [11] propose a Two-stage Multi-teacher Knowledge Distillation method for web Question Answering system. They first developed a general Q&A distillation task for student model pre-training, and then further fine-tuned the pre-trained student model with multi-teacher knowledge distillation on downstream tasks, which effectively reduced the overfitting bias in individual teacher models.

*3) Transliteration:* The paper "MATra: A Multilingual Attentive Transliteration System for Indian Scripts" [4] demonstrates a model that can perform transliteration between 4 Indian languages and English with extraordinary precision using a single transformer model as their backbone architecture. The paper "Effective Architectures for Low Resource Multilingual Named Entity Transliteration" [12] has represented a model that can transliterate between 590 languages and English using the transformers architecture.

## V. METHOD

### A. Dataset

The authors use the NEWS 2012 and 2018 datasets, the same dataset used in the MATra model, for cross-lingual transliteration between the 4 Indian languages - Hindi (Hin), Bengali (Ben), Tamil (Tam) and Kannada (Kan). The dataset is a collection of transliterated pairs between the Indian languages and English (Eng). Table I shows the size of the dataset in each transliteration pair.

| Language Pair | Number of samples |
|---|---|
| Hin-Eng | 14,826 |
| Ben-Eng | 15,972 |
| Tam-Eng | 13,376 |
| Kan-Eng | 12,563 |

TABLE I: The number of words in each transliteration pair with English (Eng) as the target.

The dataset is converted to a multilingual dataset with two formats:

1) Uni-Directional: This format has two possible types, one where all the Indian languages are the source and English is the target, another which is just the other way round.

2) Bi-Directional: A combination of two uni-directional datasets, i.e., Indic-to-English + English-to-Indic.

One can train two models for each of the two possible uni-directional datasets, i.e., one trained to transliterate from Indic to Latin script (English) and another from Latin script (English) to Indic script, which can then be chained up in a sequence to accomplish cross-lingual transliteration between any two Indian languages. However, both the uni-directional tasks are similar in nature, and hence MATra is trained using the bi-directional form of the dataset such that a single model can accomplish both tasks and be reused in the pipeline's two-step process for cross-lingual transliteration.

### B. Methodology

This method of knowledge distillation is a simple and elegant training method to simplify multi-stage inference pipelines to a single model. It uses knowledge distillation, a paradigm of training models where a student model is trained to match the outputs of a teacher model, to condense multiple steps in an inference pipeline with multiple models to a single model. Crosslingual transliteration via MATra is an example of such an inference pipeline.

*1) Teacher Model:* Generally speaking, two transformer sequence-to-sequence models, one trained to transliterate from Indic to Latin script and another from Latin to Indic script, are chained together to perform cross-lingual transliteration between any two Indian languages. In our case, MATra serves as the teacher model and is reused in both stages of the pipeline as it was trained using the bi-directional form of the dataset. It is trained with special language tokens, i.e., ⟨hindi⟩, ⟨bengali⟩, ⟨tamil⟩, ⟨kannada⟩, which replace the start of sequence (SOS) token in the decoder and prompt the model to generate transliteration in the desired language making it multilingual. A point to be noted is that the proposed method is designed to work with multiple models and not necessarily a single model like MATra, which is just a special case for this particular task.

*2) Student Encoder Distillation:* Figure 1 illustrates the pipeline for transliteration using two models, both of which are MATra. As shown, it follows a 4-step process to generate the final output, which can be mathematically represented as,

$$v_1 = f_{\theta_0}(x_s) \quad \text{Step 1: Take Indic input}$$
$$x_t = g_{\phi_0}(v_1) \quad \text{Step 2: Get Intermediate English output}$$
$$v_2 = f_{\theta_t}(x_t) \quad \text{Step 3: Pass } x_t \text{ as input to next model}$$
$$y = g_{\phi_t}(v_2) \quad \text{Step 4: Get final Indic output}$$

Where $f_{\theta_0}$ is the first model's encoder from the pipeline, parameterized by $\theta_0$ (as shown in Figure 1), $g_{\phi_0}$ is the decoder of the first model with $\phi_0$ as parameters. Similarly, $f_{\theta_t}$ and $g_{\phi_t}$ are the second model's encoder and decoder, respectively.

The final Indic output $y$ is dependent only on $v_2$ which takes 3 forward passes to get generated. Our proposed method aims at finding a shortcut to generating $v_2$ in a single forward pass given the original input $x_s$ and skipping the first model entirely. This is accomplished by training the student encoder $f_{\theta_s}$ to align itself to the same latent vector space as the teacher encoder $f_{\theta_t}$ such that,

$$f_{\theta_s}(x_s) = f_{\theta_t}(g_{\phi_o}(f_{\theta_o}(x_s)))$$

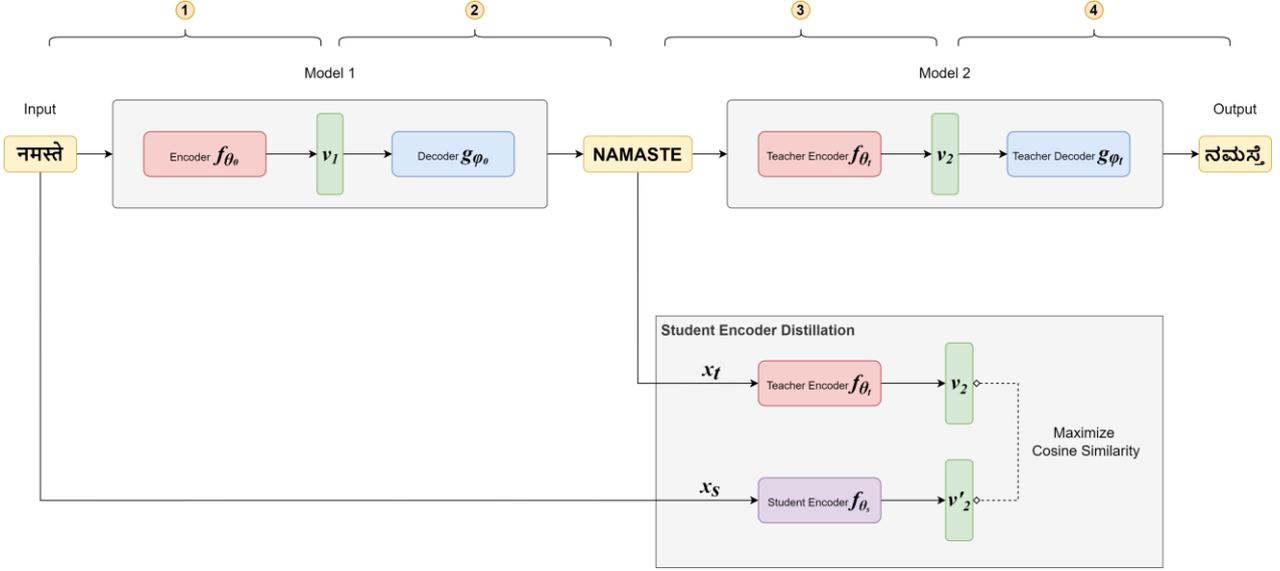

Fig. 1: The illustration here shows a pipeline of two models for crosslingual transliteration of Hindi to English to Kannada in 4 inference steps. The second model in steps 3-4 serves as the teacher set of encoder and decoder. The *Student Encoder Distillation* block illustrates the primary phase our method, which learns a direct mapping of the input $x_s$ to the encoded vector $v_2$ by entirely skipping the first model of the pipeline. The training objective is to maximize the similarity between the teacher's encoded vector $v_2$ and the student's encoded vector $v_2$.

Given that a dataset of $x_s$ and $x_t$ pairs exist, as the first model was trained on that, we pass $x_s$ and $x_t$ to the student $f_{\theta_s}$ and teacher $f_{\theta_t}$ encoders respectively. Then we measure the similarity of the output vectors from both networks using a loss function based on cosine similarity and optimize the parameters of only the student encoder while the teacher parameters are frozen. Finally, our pipeline is reduced to a single model with only 2 steps

The full student encoder training is demonstrated in the pseudo-code implementation in Algorithm 1

---

**Algorithm 1** Pseudocode for Student Encoder Training

---

**Require:** $f_{\theta_t}, f_{\theta_s}$: teacher and student encoders
**Require:** $dataloader$: first model's dataloader
**Ensure:** $\theta_t.grad = false$        ▷ Freeze teacher parameters
  **for** $x_s, x_t$ in $dataloader$ **do**
    $v_2 \leftarrow f_{\theta_t}(x_t)$        ▷ Get encoded vector from teacher
    $v_2' \leftarrow f_{\theta_s}(x_s)$        ▷ Get encoded vector from student
    $sim \leftarrow cosine\_similarity(v_2, v_2')$
    $Loss \leftarrow 1 - sim$
    $Loss.backward()$
    $Update(\theta_s)$
  **end for**

---

*3) Finetuning the Decoder:* In an ideal universe, the student encoder learns a perfect mapping from $x_s$ to $v_2$, which can be used with the teacher model's decoder to generate exact outputs as the original pipeline. However, perfectly transferring the teacher's knowledge to the student model in the encoder distillation phase is not attainable in practice. Therefore, the second phase of this knowledge distillation method focuses on adapting the teacher decoder $g_{\phi_t}$, parameterized by $\phi_t$, to the new encoded vector $v_2'$ produced by student encoder $f_{\theta_s}$. This is achieved by fine-tuning the teacher decoder $g_{\phi_t}$ with the student encoder frozen. The general implementation for decoder finetuning is demonstrated in Algorithm 2.

---

**Algorithm 2** General Pseudocode of Decoder Finetuning

---

**Require:** $g_{\phi_t}$: teacher decoder
**Require:** $f_{\theta_t}, f_{\theta_s}$: teacher and student encoders
**Require:** $dataloader$: first model's dataloader
  $g_{\phi_s} \leftarrow g_{\phi_t}$: duplicate teacher decoder as student
**Ensure:** $\phi_t.grad = false$        ▷ Freeze teacher decoder
**Ensure:** $\theta_t.grad = false$        ▷ Freeze teacher encoder
**Ensure:** $\theta_s.grad = false$        ▷ Freeze student encoder
  **for** $x_s, x_t$ in $dataloader$ **do**
    $v_2, v_2' \leftarrow f_{\theta_t}(x_t), f_{\theta_s}(x_s)$  ▷ Get the encoded vectors
    $y_t \leftarrow g_{\phi_t}(v_2)$        ▷ Decoder ground truth
    $y_s \leftarrow g_{\phi_s}(v_2')$        ▷ Predicted label
    $Loss \leftarrow loss\_function(y_t, y_s)$
    $Loss.backward()$
    $Update(\phi_s)$
  **end for**

---

In the case of MATra, the input and output languages belong to the same domain, given that it is fully cross-lingual, i.e., the pipeline can take in a Hindi word and reconstruct the same Hindi word in the output because the two stages are mirrors of each other (i.e., Indic to English and English to Indic). This special case applies to similar cross-lingual transduction model pipelines that use an intermediate. The General form of the decoder finetuning described in Algorithm 2 can be reduced to an autoencoder-like input reconstruction objective as shown in Algorithm 3

## VI. Results

We have distilled the teacher models (MATra model) into the student model (EPIK model) to demonstrate

---

**Algorithm 3** Decoder Finetuning for MATra

---

**Require:** $g_{\phi_t}$: teacher decoder
**Require:** $f_{\theta_s}$: student encoder
**Require:** *dataloader*: dataloader of inputs only
    $g_{\phi_s} \leftarrow g_{\phi_t}$: duplicate teacher decoder as student
**Ensure:** $\phi_t.grad = false$     ▷ Freeze teacher decoder
**Ensure:** $\theta_s.grad = false$     ▷ Freeze student encoder
  **for** $x_s$ in *dataloader* **do**
    $v'_2 \leftarrow f_{\theta_s}(x_s)$     ▷ Get the encoded vector
    $y_s \leftarrow g_{\phi_s}(v'_2)$     ▷ Reconstructed input
    $Loss \leftarrow loss\_function(x_s, y_s)$
    $Loss.backward()$
    $Update(\phi_s)$
  **end for**

---

knowledge transfer. Unlike the MATra model, the EPIK model can accomplish the task of transliteration between all the pairs without any intermediate English output. Thus, there is no metric to compare the absolute accuracy without the dataset. However, all other metrics, like Phonetic Accuracy and CER, are still used as benchmarks to demonstrate the performance of the EPIK model.

There is no available dataset for transliteration task between the language pairs; therefore, all the evaluations are done by crowd-sourced human evaluators. We use Phonetic Accuracy as the primary comparison metric for the EPIK model. This is primarily because transliteration is a phonetic-based task, and we have no dataset to compare the absolute accuracy. If the EPIK model incorrectly transliterates any word during the evaluation, the evaluators must write the correct word for the respective inputs. This helps in the computation of a few additional metrics like CER. In addition to the available metrics to evaluate the performance of the student model, we also compare the student model's encoder representations with the teacher model's encoder representations for each source language separately.

### A. Phonetic Accuracy

Phonetic Accuracy is defined by the number of words predicted correctly in terms of phonetics, divided by total number of words in the test dataset. This metric tells us how many words are predicted correctly and is similar to the absolute accuracy in every aspect except one. Transliteration is purely a phonetic task because there can be a single syllable written using 2 or more different letters (in any target language), and both letters are perfectly acceptable. For example, the Hindi word 'NTt' can be transliterated as 'LEAGUE' or 'LEEG' or 'LIIG', and all these words are phonetically acceptable. Thus, if absolute accuracy were used as the primary metric, several such ambiguous instances would naturally exist where the transliterated word is phonetically correct but incorrect if letter-wise accuracy is considered. Due to such discrepancies, we use phonetic accuracy instead of absolute accuracy as our primary metric.

For each of the 12 language pairs, there are 1000 words in each test dataset, and a human evaluator manually checks the predictions. Table II shows the phonetic accuracy of the EPIK model for each language pair. We also compare the phonetic accuracy of each language pair from the MATra model (teacher model) because we are distilling from the MATra model. The MATra model had an average phonetic accuracy of 93.5%, while the EPIK model has an average phonetic accuracy of 92.1%. The average reduction in phonetic accuracy is 1.44%. 5 out of the 12 pairs have a reduction of 0.1% or lesser in terms of phonetic accuracy, which implies that only 1 extra mistake was made in each dataset (out of 1000 examples tested). However, because a human has evaluated each dataset, the chances of human errors are likely, and a few of the samples of the test dataset might be incorrectly evaluated. Thus, the reduction of 0.1% (or about 1 extra mistake) for each pair could be ignored, and these pairs could be considered perfectly distilled.

| Language Pair | MATra Accuracy | EPIK Accuracy |
|---|---|---|
| Hin-Ben | 0.937 | 0.918 |
| Hin-Tam | 0.968 | 0.919 |
| Hin-Kan | 0.935 | 0.899 |
| Ben-Hin | 0.946 | 0.935 |
| Ben-Tam | 0.922 | 0.908 |
| Ben-Kan | 0.905 | 0.878 |
| Tam-Hin | 0.913 | 0.901 |
| Tam-Ben | 0.939 | 0.938 |
| Tam-Kan | 0.942 | 0.941 |
| Kan-Hin | 0.948 | 0.948 |
| Kan-Ben | 0.932 | 0.932 |
| Kan-Tam | 0.944 | 0.944 |

TABLE II: *Phonetic Accuracy (higher the better) of the MATra model and EPIK model. The average phonetic accuracy of the MATra model is 93.5%, while the average phonetic accuracy of the EPIK model is 92.1%. The net reduction in accuracy is only 1.44%.*

### B. CER (Character Error Rate)

CER indicates how many character-level deletions, insertions or substitutions are required to convert the predicted word to the ground truth. Therefore, a lower value of CER shows that the words are predicted more accurately at the character level.

While evaluating each word manually, the evaluators write the correct prediction for each incorrect word, which helps us compute such metrics. The scores are obtained from the formula -

$$CER = \frac{S + D + I}{S + D + I + C} \quad (1)$$

Where,
S = Number of Substitutions
D = Number of Deletions
I = Number of Insertions
C = Number of Correct Characters

In any language, a few word pairs might have a high CER score but sound phonetically similar. For example, the words 'know' and 'no' sound similar phonetically but have a CER score of 0.5 for this pair. Thus, this is not the ideal method to evaluate the model, especially in phonetic tasks. This is the primary reason behind the fact that even after a reduction of CER scores for a few language pairs, we do not see a high drop in phonetic accuracy.

Table III shows the CER scores of the EPIK model. In addition, we also compare the CER scores with the teacher model (MATra model) for each language pair. The average reduction in CER is only 1.04% from the teacher model. Out of the 12 pairs, 5 pairs have a reduction of only

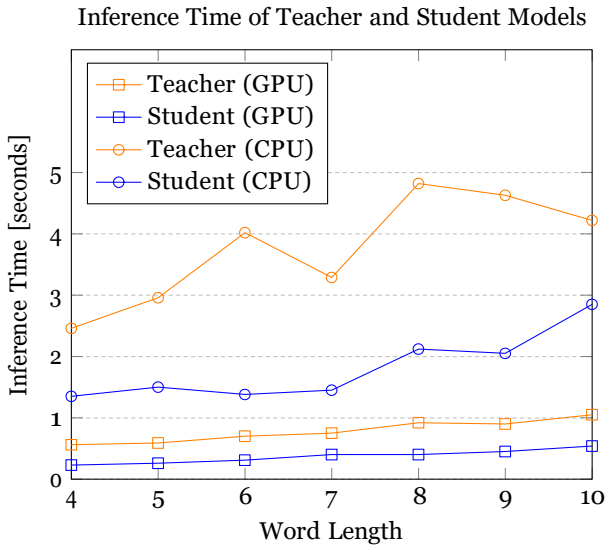

Fig. 2: Time taken for predicting the output word by the teacher (MATra) and student (EPIK MATra). The GPU used is an RTX 2060 with 6GB VRAM.

0.1% or lesser. This reduction could be due to strictness in evaluation by any evaluator (just like the reduction of phonetic accuracy), and can be neglected. Thus, these 5 language pairs can be considered perfectly distilled.

| Language Pair | MATra CER | EPIK CER |
|---|---|---|
| Hin-Ben | 0.059 | 0.072 |
| Hin-Tam | 0.086 | 0.112 |
| Hin-Kan | 0.074 | 0.099 |
| Ben-Hin | 0.054 | 0.064 |
| Ben-Tam | 0.1 | 0.115 |
| Ben-Kan | 0.08 | 0.104 |
| Tam-Hin | 0.057 | 0.068 |
| Tam-Ben | 0.079 | 0.080 |
| Tam-Kan | 0.082 | 0.082 |
| Kan-Hin | 0.036 | 0.036 |
| Kan-Ben | 0.064 | 0.064 |
| Kan-Tam | 0.086 | 0.086 |

TABLE III: *Comparison of Character Error Rate values (lower the better) between student model (EPIK model) and teacher model (MATra model).*

### C. Inference Time and Memory Consumption

The time for predicting an output by the teacher model (MATra model, shown in orange colour) and student model (EPIK model, shown in blue colour) are shown in Fig 2, on both the CPU (represented by the dotted lines) and GPU (represented by the bold lines). The measurement of this metric was done on a system with NVidia RTX 2080 GPU, Intel i7 processor with 16GB RAM. The average times of execution are calculated by taking the weighted average of execution time for the length of words, and the results of average times of running on CPU and GPU are as follows -

1) Teacher model (MATra model) on CPU: 3.15 sec

2) Student model (EPIK model) on CPU: 1.40 sec

3) Teacher model (MATra model) on GPU: 0.64 sec

4) Student model (EPIK model) on GPU: 0.30 sec

The average execution time reduction is about 55.5% on the CPU and about 53.1% on the GPU while comparing the student and teacher model. In general, the sizes and architecture of both the teacher and student model are exactly similar, and thus the expected reduction could be around 50%. However, in the teacher model (MATra model), the intermediate output is input into the model again, requiring a few extra milliseconds for processing. The extra reduction of 5.5% on CPU and 3.1% on GPU for the student model is due to the elimination of the intermediate output.

This technique of knowledge distillation can easily be implemented for multiple teacher models converted into a single student model. Therefore, the final time for prediction could be expected to reduce by a factor of 1/N (where 'N' is the number of models in the chain) as compared to the total time of execution of all the models in the chain.

### D. Encoder Similarity

Table IV shows the average similarity of all encoder representations between the teacher model (MATra model) and student model (EPIK MATra model). The encoder representation of both models is a constant-sized matrix. The similarity score is the average of Cosine Similarity scores between the encoder representation of the teacher and student model for each word in each language pair.

The average similarity of all the pairs is 97.5%. In Table IV, the similarity scores are shown for each source language for both the train and test datasets. For example, the similarity score is 96.3% for the test datasets where Hindi is the source language, which includes all 3 language pairs: Hin-Ben, Hin-Kan and Hin-Tam. The average reduction in similarity score is only 0.98% between train and test datasets, indicating no overfitting. The similarity scores of the test dataset are higher than the training dataset, wherever Kannada is the source language. This is why the phonetic accuracy is also considerably higher than other language pairs where Kannada is the source.

On average, few language pairs have similarity scores of only 97.5% or lesser, and the model could be trained for longer to maximize the similarity. However, as elaborated in section VI.E, there are a few examples where the student model can outperform the teacher model even though it has been distilled from the teacher model. This reduction of about 2.5% has helped eliminate noise from the model's prediction, reduce overfitting and predict few test data samples more accurately than the teacher model.

| Source Language | Train Data | Test Data |
|---|---|---|
| Hin | 0.975 | 0.963 |
| Ben | 0.978 | 0.958 |
| Tam | 0.971 | 0.957 |
| Kan | 0.976 | 0.982 |

TABLE IV: Encoder Similarity of encoder representations for each source language between teacher model (MATra model) and student model (EPIK model). The average similarity is 97.5%

### E. Emergent Results of Distilled Model

Few examples from the test dataset are chosen and the predictions are shown in the tables V VI VII VIII for the EPIK model. In addition, the same words are predicted using the MATra model and the predictions from the MATra model are also compared beside. The predictions from both the models for the randomly chosen examples are perfectly identical and the student model is a replica of the teacher model.

| Pair | Input Word | MATra Prediction | Distil Model Prediction |
|---|---|---|---|
| Hin-Ben | अनवर | আনওয়ার | আনওয়ার |
| Hin-Ben | गोडेल | গোডেল | গোডেল |
| Hin-Tam | आसही | அசாஹி | அசாஹி |
| Hin-Tam | केनन | கென்னன் | கென்னன் |
| Hin-Kan | मुईज़ | ಮುಯಿಜ್ | ಮುಯಿಜ್ |
| Hin-Kan | आसिफ | ಆಸಿಫ್ | ಆಸಿಫ್ |

TABLE V: Few examples of transliterated outputs from the EPIK model, for all language pairs with source as Hindi. The predictions of the teacher model (MATra model) are also compared for reference.

| Pair | Input Word | MATra Prediction | Distil Model Prediction |
|---|---|---|---|
| Ben-Hin | রামফল | रामफल | रामफल |
| Ben-Hin | ধরমবীর | धरमबीर | धरमबीर |
| Ben-Tam | হোমে | ஹோமாம்மே | ஹோமாம்மே |
| Ben-Tam | মৌসম | மௌசம் | மௌசம் |
| Ben-Kan | অলকা | ಅಲೋಕಾ | ಅಲೋಕಾ |
| Ben-Kan | অনীক | ಅನೀಕ್ | ಅನೀಕ್ |

TABLE VI: Few examples of transliterated outputs from the EPIK model, for all language pairs with source as Bengali. The predictions of the teacher model (MATra model) are also compared for reference.

| Pair | Input Word | MATra Prediction | Distil Model Prediction |
|---|---|---|---|
| Kan-Hin | ಲಠಕೆ | लड़की | लड़की |
| Kan-Hin | ಕಿರಣ | किरणा | किरणा |
| Kan-Ben | ಅಂಗರೆ | আঙ্গারা | আঙ্গারা |
| Kan-Ben | ಸಫ್ಫರ್ | সাফার | সাফার |
| Kan-Tam | ಅಚಲಾ | அச்சலா | அச்சலா |
| Kan-Tam | ಆಲ್ಹಾ | ஆல்ஹ்ஹா | ஆல்ஹ்ஹா |

TABLE VII: Few examples of transliterated outputs from the EPIK model, for all language pairs with source as Kannada. The predictions of the teacher model (MATra model) are also compared for reference.

| Pair | Input Word | MATra Prediction | Distil Model Prediction |
|---|---|---|---|
| Tam-Hin | தைடேவ | तायपेई | तायपेई |
| Tam-Hin | கஹாணி | कहानी | कहानी |
| Tam-Ben | கரலே | কারালে | কারালে |
| Tam-Ben | அசோக | অশোক | অশোক |
| Tam-Kan | ஹ்ருயே | ಹುಯ್ | ಹುಯ್ |
| Tam-Kan | அலோஸாம் | ಅಸೋಸಾಮ್ | ಅಸೋಸಾಮ್ |

TABLE VIII: Few examples of transliterated outputs from the EPIK model, for all language pairs with source as Tamil with the predictions of the teacher model (MATra model).

The student model (EPIK model) has been distilled from the teacher model (MATra model) and is able to replicate the teacher model in almost every example. However, there are few instances where the student model outperforms the teacher model giving more precise results. The average encoder similarity between the student and teacher model is only 97.5%. This reduction of about 2.5% helped the student model remove excess noise and avoid overfitting.

In addition to the general predictions, Tables IX X XI XII show few predictions for each language pair where the student model has outperformed the teacher model (the student model has predicted the correct output while the teacher model has made an error).

| Pair | Input Word | MATra Prediction | Distil Model Prediction |
|---|---|---|---|
| Hin-Ben | आएँगी | আংগী | আয়েঙ্গি |
| Hin-Ben | बिटर | থিরেট | থিমেটার |
| Hin-Tam | रशीद | ரவீ | ரவ்ஷீத் |
| Hin-Tam | हावर्ड | ஹவ்கர் | ஹார்வர்ட் |
| Hin-Kan | अशरफ | ಅಶ್ರಣ | ಅಶ್ರಫ್ |
| Hin-Kan | सिमंस | ಸಿಮ್ಎ | ಸಿಮ್ಮನ್ಸ್ |

TABLE IX: Few examples of transliterated outputs with source as Hindi, where student model (EPIK model) outperformed teacher model (MATra model). The words highlighted are correct while the non-highlighted words are incorrect.

| Pair | Input Word | MATra Prediction | Distil Model Prediction |
|---|---|---|---|
| Ben-Hin | ল্যাণ্ড | लैन्ड्स | लैंड |
| Ben-Hin | ইনশুলাব | इंखुबा | इंकलाब |
| Ben-Tam | ঋষিকেশ | ஹிரிர்ஹிகேஷ் | ரிஷிகேஷ் |
| Ben-Tam | আসাম | அஸமா | அஸாம் |
| Ben-Kan | শিলিং | ಶೆಲಿನ | ಶೆಲಿಂಗ್ |
| Ben-Kan | হোপস | ಹಾಪ್ | ಹೋಪ್ಸ್ |

TABLE X: Few examples of transliterated outputs with source as Bengali, where student model (EPIK model) outperformed teacher model (MATra model). The words highlighted are correct while the non-highlighted words are incorrect.

| Pair | Input Word | MATra Prediction | Distil Model Prediction |
|---|---|---|---|
| Kan-Hin | ಮಿಯಾಚ್ | मिआक्क | मिआच |
| Kan-Hin | ಕಟ್ಲರ್ | कटलर्स | कटलर |
| Kan-Ben | ಅನಂಗ | অনং | আনাঙ্গা |
| Kan-Ben | ಸುಗರ್ | শগ | সুগার |
| Kan-Tam | ಎಕ್ಷಂ | விஷ்ம் | விஷ்யம் |
| Kan-Tam | ಅಟ್ಟುನೆಟಿ | அட்டுனிட் | அட்டனிட்டி |

TABLE XI: Few examples of transliterated outputs with source as Kannada, where student model (EPIK model) outperformed teacher model (MATra model). The words highlighted are correct.

## VII. LIMITATIONS

The model(s) trained in this format are able to replicate the teacher model completely, and the student model is able to capture all the errors of the teacher model as well. This technique is perfectly applicable where the teacher model is able to perform the required task with minimal or absolutely no errors. However, if we do not have a teacher model that gives accurate results, the student model will perform poorly at every place where the teacher model makes an error.

| Pair | Input Word | MATra Prediction | Distil Model Prediction |
|------|-----------|------------------|-------------------------|
| Tam-Hin | கின் | किन्द | किन |
| Tam-Hin | அச்சல் | अचास | अचल |
| Tam-Ben | ரெயின் | রাইন | রাইন |
| Tam-Ben | ஸ்வொன் | সোয়াস্ | সোয়ান |
| Tam-Kan | ஷகரீ | ಶ್ಕೇರೀ | ಶ್ಕೇ |
| Tam-Kan | சத்தே | ಚಾತ್ತ್ | ಸತ್ತೆ |

TABLE XII: Few examples of transliterated outputs with source as Tamil, where student model (EPIK model) outperformed teacher model (MATra model). The words highlighted are correct while the non-highlighted words are incorrect.

Furthermore, because there is no direct dataset available, computer-based evaluation is not possible. To check the performance of such models, human evaluation is always required.

## VIII. FUTURE WORK

The student model replicates the teacher model(s), and carries along almost all the errors as well. A new neural network-based feedback loop can be implemented to alert the encoder or decoder while training to understand the difference between correct and incorrect predictions. Because there is no dataset for the student model, we can not directly separate the correct and incorrect examples beforehand. Instead, we need to apply a neural network-based feedback loop that can separate the incorrect examples dynamically. This way, the errors of the teacher model(s) could be reduced or eliminated.

Regarding the distillation of MATra, the size and shape (same number of weights and layers) of both teacher and student models are exactly similar, and thus, the size of the EPIK model is also exactly similar to the MATra model. We could generate another student model that can be distilled in terms of size additionally, instead of knowledge only.

## IX. CONCLUSION

The EPIK model is distilled from the MATra model using a completely novel method of knowledge distillation. The distillation process works without the existence of direct datasets, and primarily focuses on the distillation of knowledge from multiple teacher models to a single student model. This distillation technique is unconventional in the task of knowledge distillation where direct datasets do not exist, and currently, models perform the required tasks in a chaining mechanism. This method can also be used to distil knowledge between other encoder-decoder-based architectures as well, like LSTMs and Transformers.

The EPIK model can execute direct transliteration between any pair between four languages, where no direct datasets exist. The average phonetic accuracy is 92.1% (transliteration is primarily a phonetic-based task) and the average CER score is 0.015. The time for execution has been reduced by 54.3%, and the EPIK model (student model) is able to outperform the MATra model (teacher model) in a few examples even though it has been distilled from the MATra model (teacher model).